\documentclass{article}
\usepackage{amsmath,amsfonts}
\usepackage{mathtools}
\usepackage{todonotes}
\usepackage{graphicx} % more modern

\usepackage{subfigure}
\graphicspath{ {./} }
\usepackage{float}
\usepackage{natbib}
\usepackage{hyphenat}

\usepackage{algorithm}
\usepackage{algorithmic}

\newcommand{\floorx}[1]{\left \lfloor #1 \right \rfloor} 
\newcommand{\fxpt}[2]{$\tt {\left<{#1},{#2} \right>}$}
 
\usepackage[accepted]{icml2015}

\begin{document} 

\twocolumn[
\icmltitle{Deep Learning with Limited Numerical Precision}

% unless you've provided the [accepted] option to the icml2015
\icmlauthor{Suyog Gupta}{suyog@us.ibm.com}
\icmlauthor{Ankur Agrawal}{ankuragr@us.ibm.com}
\icmlauthor{Kailash Gopalakrishnan}{kailash@us.ibm.com}

\icmladdress{IBM T. J. Watson Research Center,
            Yorktown Heights, NY 10598}
\icmlauthor{Pritish Narayanan}{pnaraya@us.ibm.com}
\icmladdress{IBM Almaden Research Center,
            San Jose, CA 95120}

\icmlkeywords{Limited Precision Computation, Stochastic Rounding, Training Deep Neural Networks}

\vskip 0.3in
]

\begin{abstract} 
Training of large-scale deep neural networks is often constrained by the available computational resources.~We study the effect of limited precision data representation and computation on neural network training.~Within the context of low-precision fixed-point computations, we observe the rounding scheme to play a crucial role in determining the network's behavior during training.~Our results show that deep networks can be trained using only $16$-bit wide fixed-point number representation when using stochastic rounding, and incur little to no degradation in the classification accuracy.~We also demonstrate an energy-efficient hardware accelerator that implements low-precision fixed-point arithmetic with stochastic rounding.
%for prototyping, an implementation of the stochastic rounding scheme that minimizes the hardware overhead while achieving high throughput for fixed-point computations.
\end{abstract} 

\section{Introduction}

\label{Introduction}
To a large extent, the success of deep learning techniques is contingent upon the underlying hardware platform's ability to perform fast, supervised training of complex networks using large quantities of labeled data.~Such a capability enables rapid evaluation of different network architectures and a thorough search over the space of model hyperparameters.~It should therefore come as no surprise that recent years have seen a resurgence of interest in deploying large-scale computing infrastructure designed specifically for training deep neural networks. Some notable efforts in this direction include distributed computing infrastructure using thousands of CPU cores \cite{dean2012large, microsoftAdam}, or high-end graphics processors (GPUs) \cite{krizhevsky2009learning}, or a combination of CPUs and GPUs scaled-up to multiple nodes \cite{coates2013deep,wu2015deep}.
% a capable computing infrastructure

At the same time, the natural error resiliency of neural network architectures and learning algorithms is well-documented, setting them apart from more traditional workloads that typically require precise computations and number representations with high dynamic range.~It is well appreciated that in the presence of statistical approximation and estimation errors, high-precision computation in the context of learning is rather unnecessary \citep{BottouBousquet}.~Moreover, the addition of noise during training has been shown to improve the neural network's performance \cite{murray94,bishop95,audhkhasi2013noise}.~With the exception of employing the asynchronous version of the stochastic gradient descent algorithm \cite{recht2011hogwild} to reduce network traffic, the state-of-the-art large-scale deep learning systems fail to adequately capitalize on the error-resiliency of their workloads.~These systems are built by assembling general-purpose computing hardware designed to cater to the needs of more traditional workloads, incurring high and often unnecessary overhead in the required computational resources.

\hyphenpenalty=100
The work presented in this paper owes its inception to the thinking that it may be possible to leverage algorithm-level noise-tolerance to relax certain constraints on the underlying hardware, leading to a hardware-software co-optimized system that achieves significant improvement in computational performance and energy efficiency.~Allowing the low-level hardware components to perform approximate, possibly non-deterministic computations and exposing these hardware-generated errors up to the algorithm level of the computing stack forms a key ingredient in developing such systems.~Additionally, the low-level hardware changes need to be introduced in a manner that preserves the programming model so that the benefits can be readily absorbed at the application-level without incurring significant software redevelopment costs.

\hyphenpenalty=750
As a first step towards achieving this cross-layer co-design, we explore the use of low-precision fixed-point arithmetic for deep neural network training with a special focus on the rounding mode adopted while performing operations on fixed-point numbers.~The motivation to move to fixed-point arithmetic (from the conventional floating-point computations) is two-fold.~Firstly, fixed-point compute units are typically faster and consume far less hardware resources and power than floating-point engines.~The smaller logic footprint of the fixed-point arithmetic circuits would allow for the instantiation of many more such units for a given area and power budget.~Secondly, low-precision data representation reduces the memory footprint, enabling larger models to fit within the given memory capacity.~Cumulatively, this could provide dramatically improved data-level parallelism.

The key finding of our exploration is that deep neural networks can be trained using low-precision fixed-point arithmetic, \textit{provided that the stochastic rounding scheme is applied while operating on fixed-point numbers}.~We test the validity of the proposed approach by training deep neural networks for the MNIST and CIFAR10 image classification tasks.~Deep networks trained using $16$-bit wide fixed-point and stochastic rounding achieve nearly the same performance as that obtained when trained using 32-bit floating-point computations.
Furthermore, we present a hardware accelerator design, prototyped on an FPGA, that achieves high throughput and low power using a large number of fixed-point arithmetic units, a dataflow architecture, and compact stochastic rounding modules.
% In addition, we also present a hardware design for high-throughput matrix multiplication that implements the stochastic rounding with less than 4\% overhead with respect to usage of hard-coded digital signal processing (DSP) units in the FPGA.

\section{Related Work}
Determining the precision of the data representation and the compute units is a critical design choice in the hardware (analog or digital) implementation of artificial neural networks.~Not surprisingly, a rich body of literature exists that aims to quantify the effect of this choice on the network's performance. However, a disproportionately large majority of these studies are focused primarily on implementing just the feed-forward (inference) stage, assuming that the network is trained offline using high precision computations.~Some recent studies that embrace this approach have relied on the processor's vector instructions to perform multiple $8$ bit operations in parallel~\cite{vanhoucke2011improving}, or employ reconfigurable hardware (FPGAs) for high-throughput, energy-efficient inference~\cite{ farabet2011neuflow, gokhale2014240}, or take the route of custom hardware implementations  \cite {kim2014x1000,merolla2014million}.

Previous studies have also investigated neural network training using different number representations.~Iwata \textit{et al}. \cite{iwata1989artificial} implements the back-propagation algorithm using $24$-bit floating-point processing units. Hammerstrom ~\cite{hammerstrom1990vlsi} presents a framework for on-chip learning using $8$ to $16$ bit fixed-point arithmetic.~In \cite{holi1993finite}, the authors perform theoretical analysis to understand a neural network's ability to learn when trained in a limited precision setting. Results from empirical evaluation of simple networks indicate that in most cases, $8$-$16$ bits of precision is sufficient for back-propagation learning.~In~\cite{hohfeld1992probabilistic}, probabilistic rounding of weight updates is used to further reduce ($<$ 8 bits) the precision requirements in gradient-based learning techniques.~While these studies provide valuable insights into the behavior of the limited precision training of neural networks, the networks considered are often limited to variants of the classical multilayer perceptron containing a single hidden layer and only a few hidden units.~Extrapolating these results to the state-of-the-art deep neural networks that can easily contain millions of trainable parameters is non-trivial.~Consequently, there is a need to reassess the impact of limited precision computations within the context of more contemporary deep neural network architectures, datasets, and training procedures. 

A recent work \cite{chen2014dadiannao} presents a hardware accelerator for deep neural network training that employs fixed-point computation units, but finds it necessary to use $32$-bit fixed-point representation to achieve convergence while training a convolutional neural network on the MNIST dataset.~In contrast, our results show that it is possible to train these networks using only $16$-bit fixed-point numbers, so long as stochastic rounding is used during fixed-point computations.~To our knowledge, this work represents the first study of application of stochastic rounding while training deep neural networks using low-precision fixed-point arithmetic.
\section{Limited Precision Arithmetic}
Standard implementations of deep neural network training via the back-propagation algorithm typically use 32-bit floating-point ({\tt float}) representation of real numbers for data storage and manipulation. 
%The {\tt float} datatype divides the 32 bits into the sign-exponent-mantissa as $1$-$8$-$23$, delivering an extremely large dynamic range ?? and ??. This may seem somewhat wasteful, especially when considering the observation that the neural network parameters (layer weights and biases) tend to be confined in a small range around zero (typically $[-1,1]$). In addition, the layer outputs and the back-propagated errors also tend to show a limited dynamic range.{\footnote{Although rectified linear units (ReLU) have unbounded activation functions, our results presented in later sections indicate that clipping the ReLU function at some reasonably high positive value goes unnoticed}.
Instead, consider the generalized fixed-point number representation: {\tt $\left[\textbf{QI}.\textbf{QF}\right]$}, where {\tt \textbf{QI}} and {\tt \textbf{QF}} correspond to the integer and the fractional part of the number, respectively.~The number of integer bits (${\tt {IL}}$) plus the number of fractional bits (${\tt {FL}}$) yields the total number of bits used to represent the number. The sum {$ \tt {IL}$ +  $\tt {FL} $} is referred to as the word length ${\tt {WL}}$.~In this paper, we use the notation \fxpt{IL}{FL} to denote a fixed-point representation in which {$ \tt {IL}$} (${\tt {FL}}$) correspond to the length of the integer (fractional) part of the number.~We also employ $\epsilon$ to denote the smallest positive number that may be represented in the given fixed-point format.~Therefore, the \fxpt{IL}{FL} fixed-point format limits the precision to $\tt {FL}$ bits, sets the range to $\left[-2^{{\tt{IL}}-1} , 2^{{\tt{IL}}-1} - 2^{-{\tt{FL}}}\right]$, and defines $\epsilon$ to be equal to $2^{-{\tt{FL}}}$.

\subsection{Rounding Modes}
As will be evident in the sections to follow, the rounding mode adopted while converting a number (presumably represented using the {\tt float} or a higher precision\footnote{We call \fxpt{IL_1}{FL_1} to be a higher precision representation than \fxpt{IL_2}{FL_2} iff $\tt {FL_1} > \tt {FL_2}$ } fixed-point format) into a lower precision fixed-point representation turns out to be a matter of important consideration while performing computations on fixed-point numbers. Given a number $x$ and the target fixed-point representation \fxpt{IL}{FL}, we define $\floorx{x}$ as the largest integer multiple of  $\epsilon ~(= 2^ {\tt{-FL}})$ less than or equal to $x$ and consider the following rounding schemes:
\setlength{\leftmargini}{1.0em}
\begin{itemize}
%\begin{comment}
%\item
%Truncation
%\begin{align*}
%Round(x,{\tt {\left<{IL},{FL} \right>}}) = \left \lfloor {x} \right \rfloor
%\end{align*}
%This scheme can be implemented quite trivially in hardware by discarding the bits in $x$ that are at a precision lower than $2^ {\tt{-FL}}$.
%\end{comment}
\item
Round-to-nearest
\begin{align*}
\begin{split}
 Round&(x,{\tt {\left<{IL},{FL} \right>}}) = \\
  &\begin{dcases}
   \floorx{x} & \text{if } \floorx{x} \leq x \leq \floorx{x} + \frac{\epsilon}{2} \\
   \floorx{x} + \epsilon & \text{if } \floorx{x} + \frac{\epsilon}{2} < x \leq \floorx{x} + \epsilon
  \end{dcases}
  \end{split}
\end{align*}
\item
Stochastic rounding: The probability of rounding $x$ to $\floorx{x}$ is proportional to the proximity of $x$ to $\floorx{x}$:
\[
 Round\left(x,{\tt {\left<{IL},{FL} \right>}}\right) =
  \begin{dcases}
   \floorx{x} & \text{w.p. } 1- \frac{x - \floorx{x}}{\epsilon} \\
   \floorx{x} + \epsilon & \text{w.p. } \frac{x - \floorx{x}}{\epsilon} \\ 
  \end{dcases}
\]
Stochastic rounding is an unbiased rounding scheme and possesses the desirable property that the expected rounding error is zero, i.e.
$\mathbb{E} \left( Round\left( x,{\tt {\left<{IL},{FL} \right>}}\right)\right) = x$

\end{itemize}
Irrespective of the rounding mode used, if $x$ lies outside the range of \fxpt{IL}{FL}, we saturate the result to either the lower or the upper limit of \fxpt{IL}{FL}:
\setlength{\leftmargini}{1.0em}
\begin{equation}
\begin{split}
Con&vert\left(x,{\tt {\left<{IL},{FL} \right>}}\right) = \\
 & \begin{dcases}
   -2^{{\tt{IL}}-1} & \text{if } x \leq -2^{{\tt{IL}}-1} \\
	2^{{\tt{IL}}-1} - 2^{-{\tt{FL}}} & \text{if } x \geq 2^{{\tt{IL}}-1}- 2^{-{\tt{FL}}}\\
   Round(x,{\tt {\left<{IL},{FL} \right>}}) & \text{otherwise} 
  \end{dcases}
\end{split}
\label{convert}
  \end{equation}

\subsection{Multiply and accumulate ({\tt {\textbf{MACC}}}) operation }
Consider two $d$-dimensional vectors $\textbf{a}$ and $\textbf{b}$ such that each component is represented in the fixed-point format \fxpt{IL}{FL}, and define $c_0 = \textbf{a}.\textbf{b}$ as the inner product of $\textbf{a}$ and $\textbf{b}$. $c_0$ is also represented in some fixed-point format {$\tt {\left<{{\tilde{IL}}},{{\tilde{IF}}} \right>}$}. We split the computation of $c_0$ into the following two steps:
\begin{enumerate}
\item
Compute $z = \sum_{i = 1}^{d}a_ib_i$

The product of $a_i$ and $b_i$ produces a fixed-point number in the \fxpt{2*IL}{2*FL} format.
$z$ can be thought of as a temporary fixed-point register with enough width (number of bits) to prevent saturation/overflow and avoid any loss of precision while accumulating the sum over all products $a_ib_i$. The requirement on the width of $z$ is $log_2d + 2\tt{WL}$ in the worst case. Note that the worst case is extremely rare and occurs when \textit{all} $a_i$ and $b_i$ are saturated to either the lower or the upper limit of  \fxpt{IL}{FL}.
\item
Convert: $c_0 = Convert(z,{\tt{\left<{{\tilde{IL}}},{{\tilde{IF}}} \right>}})$

This step invokes the $Convert()$ function defined previously in eq.~\ref{convert}, resulting in either \textit{clipping} the value in $z$ to the limits set by {$\tt {\left<{{\tilde{IL}}},{{\tilde{IF}}} \right>}$} or rounding to $\tt{\tilde{FL}}$ bits of fractional precision using the specified rounding mode. 
\end{enumerate} 
 \begin{figure*}[ht]
\vskip -0.1in
\begin{center}

\subfigure{
	\includegraphics[width = 6in]{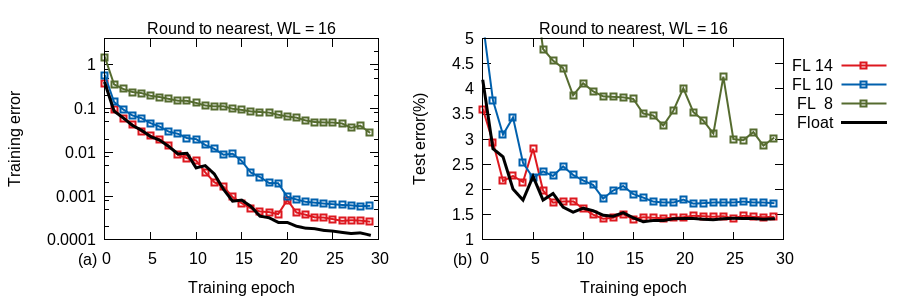}
}

\vskip -0.05in
\subfigure{
	\includegraphics[width = 6in]{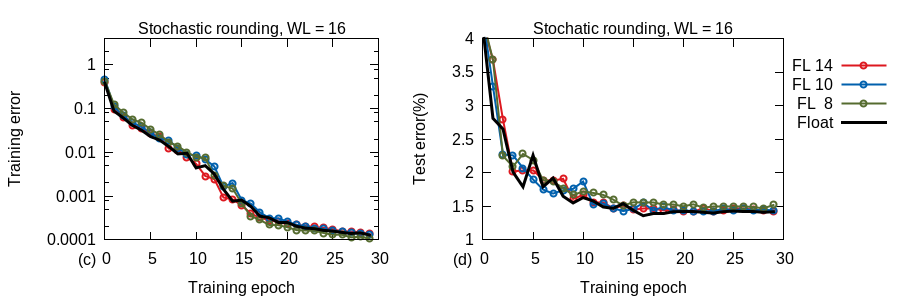}
	}
\vskip -0.1in
\caption{\textit{MNIST dataset using fully connected DNNs:} Training error (\textit{a, c}) and the test error (\textit{b, d}) for training using fixed-point number representation and rounding mode set to either ``Round to nearest'' (\textit{top}) or ``Stochastic rounding'' (\textit{bottom}). The word length for fixed-point numbers $\tt WL$ is kept fixed at 16 bits and results are shown for three different fractional (integer) lengths: 8(8), 10(6), and 14(2) bits. Results using {\tt float} are also shown for comparison. }
\label{MNIST_DNN_NormRound}
\end{center}
\vskip -0.1in
\end{figure*}
Adopting this two-step approach has several advantages.~Firstly, it closely mimics the behavior of the hardware implementation of vector inner product using the the hardware DSP\footnote{\textbf{D}igital \textbf{S}ignal \textbf{P}rocessing units are hardware units in the FPGA fabric that implement fixed-point multiplication and addition} units in FPGAs.~These DSP units accept $18$-bit inputs and accumulate the results of the {\tt {MACC}} operation in a $48$-bit wide register.~Secondly, by invoking the rounding mode only after the accumulation of all the sums, we significantly reduce the hardware overhead in implementing the stochastic rounding scheme.~Lastly, the adoption of this approach allows us to efficiently simulate fixed-point computations using CPUs/GPUs and vendor-supplied BLAS\footnote{\textbf{B}asic \textbf{L}inear \textbf{A}lgebra \textbf{S}ubprograms} libraries.~For instance, matrix multiplication of two fixed-point matrices $A$ and $B$ can be simulated by first converting them into {\tt float} matrices, calling the hardware-optimized {\tt SGEMM} routine and applying the $Convert()$ function to each element of the resulting {\tt float} matrix.

\section{Training Deep Networks}
\begin{figure*}[ht]
\vskip -0.1in
\center
\includegraphics[width=6.8in]{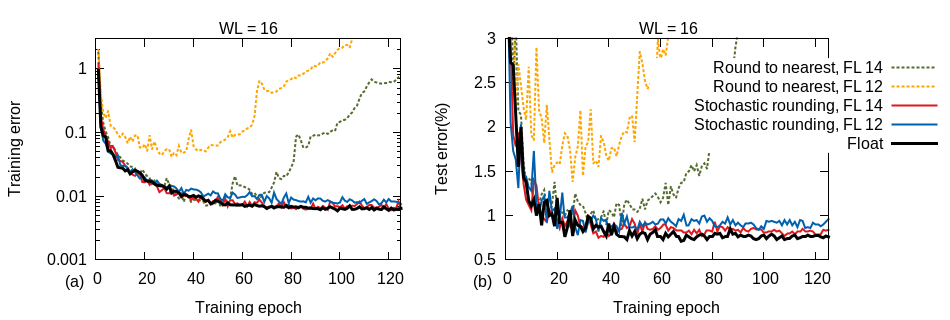}
\vskip -0.1in
\caption{\textit{MNIST dataset using CNNs:} Training error (\textit{a}) and the test error (\textit{b}) for training using fixed-point number representation and rounding mode set to either ``Round to nearest'' or ``Stochastic rounding''. The word length for fixed-point numbers $\tt WL$ is kept fixed at 16 bits and results are shown for different fractional (integer) lengths for weights and weight updates: $12(4)$, and $14(2)$ bits. Layer outputs use \fxpt{6}{10} format in all cases. Results using {\tt float} are also shown for comparison. }

\label{MNIST_CNN}
\end{figure*}

In this section, we present the results of our investigation into the effect of employing limited precision data representation during the training of deep neural networks.~We consider both fully connected deep neural networks (DNN) as well as convolutional neural networks (CNN) and present results for the MNIST\cite{mnistlecun} and the \mbox{CIFAR10}\citep{krizhevsky2009learning} datasets. As a baseline for comparison, we first evaluate the network performance (in terms of the rate of reduction of both the training error and the error on the test set) using the conventional $32$-bit floating-point arithmetic.~Subsequently, we constrain the neural network parameters (weights $W^{l}$, biases $B^{l}$), as well as the other intermediate variables generated during the back-propagation algorithm (layer outputs $Y^l$, back-propagated error $\delta^{l}$, weight updates $\Delta W^{l}$, bias updates $\Delta B^{l}$) to be represented in the fixed-point format and train the network again starting from random initialization of the parameters.~While training using fixed-point, the different model hyperparameters such as weight initialization, regularization parameters, learning rates etc. are kept unchanged from the ones used during the baseline evaluation.~The word length ${\tt WL}$ for the fixed-point format is set to $16$ bits i.e. the number of bits allocated to represent the integer and the fractional parts add up to $16$.  

This fairly restrictive choice of number representation has some important implications.~From the perspective of neural network training, an aggressive reduction of the precision with which the parameter updates are computed and stored  may result in the loss of the gradient information if the updates are significantly smaller than the $\epsilon$ for the given fixed-point format.~As a consequence, this may impede the progress of the gradient descent algorithm, or worse, introduce instabilities during the training procedure.~Note that in the round-to-nearest scheme, any parameter update in the range $\left(-\frac{\epsilon}{2},\frac{\epsilon}{2}\right)$ is always rounded to zero, as opposed to the stochastic rounding scheme which maintains a non-zero probability of small parameter updates to round to $\pm\epsilon$.
Secondly, since the fixed-point format offers only a limited range, outputs of the ReLU activation function may get clipped to the upper limit set by \fxpt{IL}{FL}.~From a hardware perspective, the use of $16$-bits for data storage (instead of {\tt float}) corresponds to a factor $2$ reduction in the amount of memory needed for training a given network.~Moreover, the use of the same word length for all network variables carries with it the added advantage of simplifying the hardware implementation.
\subsection{MNIST}

\subsubsection{Fully connected DNN}

In the first set of experiments, we construct a fully connected neural network with $2$ hidden layers, each containing $1000$ units with ReLU activation function and train this network to recognize the handwritten digits from the MNIST dataset.~This dataset comprises of $60,000$ training images and $10,000$ test images -- each image is $28$ x $28$ pixels containing a digit from $0$ to $9$.~The pixel values are normalized to lie in the $[0,1]$ range.~No other form of data pre-processing or augmentation is performed.~The weights in each layer are initialized by sampling random values from $\mathcal{N}\left(0,0.01\right)$ while the bias vectors are initialized to $\textbf{0}$.~The network is trained using minibatch stochastic gradient descent (SGD) with a minibatch size of $100$ to minimize the cross entropy objective function.~The $\tt float $ baseline achieves a test error of $1.4\%$. 

Next, we retrain the network using fixed-point computations and set ${\tt WL}$ to $16$ bits.~Figure~\ref{MNIST_DNN_NormRound} shows the results for the two rounding modes: Round-to-nearest and Stochastic rounding.~In both cases, allocating $14$ bits to the fractional part\footnote {Using up $14$ bits for the fractional part leaves only $2$ bits (including the sign bit) for representing the integer portion of the number.~This does not seem to adversely affect the network performance. } produces no noticeable degradation in either the convergence rate or the classification accuracy.~A reduction in the precision below $14$ bits begins to negatively impact the network's ability to learn when the round-to-nearest scheme is adopted.~This is primarily because at reduced fractional precision, most of the parameter updates are rounded down to zero.~In contrast, the stochastic rounding preserves the gradient information, atleast statistically, and the network is able to learn with as few as $8$ bits of precision without any significant loss in performance.~Note, however, at a precision lower than $8$ bits, even the stochastic rounding scheme is unable to fully prevent the loss of gradient information.

\begin{figure*}[ht]
\vskip -0.1in
\center
\includegraphics[width=6.8in]{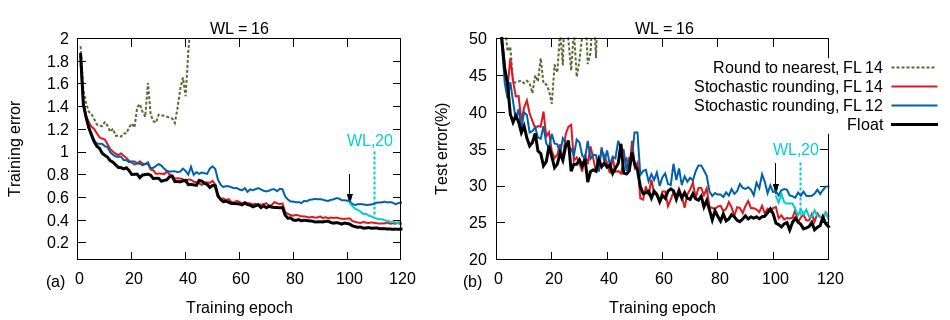}
\vskip -0.1in
\caption{\textit{CIFAR10 dataset using CNNs:}Training error (\textit{a}) and the test error (\textit{b}) for training using fixed-point number representation and rounding mode set to either ``Round to nearest'' or ``Stochastic rounding''.~The word length for fixed-point numbers $\tt WL$ is kept fixed at 16 bits and results are shown for different fractional (integer) lengths for weights and weight updates: $12(4)$, and $14(2)$ bits.~The black arrows indicate the epoch after which the training is carried out using ${\tt WL} $ = $20$ bits.~Results using {\tt float} are also shown for comparison.}
\label{CIFAR10_CNN}

\vskip -0.1in
\end{figure*} 

\subsubsection{CNN}
Using the MNIST dataset,~we also evaluate a CNN with an architecture similar to LeNet-5 \cite{lecun1998gradient}.~It comprises of $2$ convolutional layers with $5$x$5$ filters and ReLU activation function.~The first layer has $8$ feature maps while the second convolutional layer produces $16$ feature maps.~Each convolutional layer is followed by a pooling/subsampling layer.~The pooling layers implement the max pooling function over non-overlapping pooling windows of size $2$x$2$.~The output of the second pooling layer feeds into a fully connected layer consisting of $128$ ReLU neurons, which is then connected into a $10$-way softmax output layer. 

For training this network, we adopt an exponentially decreasing learning rate -- scaling it by a factor of 0.95 after every epoch of training.~The learning rate for the first epoch is set to 0.1.~Momentum ($p = 0.9$) is used to speed up SGD convergence.~The weight decay parameter is set to $0.0005$ for all layers.~When trained using {\tt float}, the network achieves a test error of $0.77\%$.~As was done previously for DNNs, we retrain the network using fixed-point computations with ${\tt{WL}}$ set to $16$ bits.~However, in this case, saturating the output of the convolutional layers to a low integer value created some difficulty in jump-starting the training procedure.~As a result, we increase the number of bits allocated for the integer part at the expense of reducing the precision and choose the \fxpt{6}{10} format for representing the layer outputs.~Figure~\ref{MNIST_CNN} compiles the results obtained using the two different rounding modes.~Unlike in the case of DNNs, when the round-to-nearest scheme is adopted during fixed-point computations, the training procedure fails to converge.~When stochastic rounding is used, we achieve a test error of $0.83\%$ and $0.90\%$ for $14$-bit and $12$-bit precision, respectively -- corresponding to only a slight degradation from the {\tt float} baseline. 

%For stochastic rounding and precision of 12 bits, we obtain a test error of $0.90\%$. However, increasing the precision to 14 bits reduces the test error to $0.83\% $ -- 

\subsection{CIFAR10}
To further test the validity of the stochastic rounding approach, we consider another commonly used image classification benchmark: CIFAR10.~The training set consists of $50,000$ RGB images of size $32$x$32$ pixels. The images are divided into $10$ classes, each containing $5,000$ images.~The test set has $10,000$ images.~We scale the image RGB values to [0,1] range and do not perform any other form of data pre-processing or augmentation.~For this dataset, we construct a CNN with $3$ convolutional layers each followed by a subsampling/pooling layer.~The convolutional layers consist of $64$ $5$x$5$ filters and the subsampling layers implement the max pooling function over a window of size $3$x$3$ using a stride of $2$.~The $3^{rd}$ pooling layer connects to a $10$-way softmax output layer. This architecture is similar to the one introduced in \cite{hinton12dropout} with the exception that it does not implement local response normalization or dropout layers. 

The network training starts off with a learning rate of $0.01$ and reduced by a factor of $2$ after $50$, $75$, and $100$ epochs.~Using $32$-bit floating point numbers for training, this network configuration misclassifies approximately $24.6\%$ of the images in the test set.~This serves as the baseline for comparing the results obtained while training the network using fixed-point computations.~Similar to earlier experiments, we set the ${\tt {WL}}$ for fixed-point number to $16$ and test the different rounding modes and fractional precision.~The layer outputs are represented in the \fxpt{4}{12} format.~ 
As observed previously and as shown in Figure~\ref{CIFAR10_CNN}, training using fixed-point with round-to-nearest scheme begins to collapse after only a few epochs.~On the contrary, the stochastic rounding scheme appears to bestow upon the training procedure a significantly higher degree of stability.~For $14$ bits of fractional precision and the stochastic rounding scheme, the network's behavior is quite similar to that observed during the baseline evaluation and achieves a test error of $25.4\%$. 

If the precision is reduced further (to $12$ bits) the convergence rate degrades as the learning proceeds and after a point, SGD stops making progress.~This is expected since at reduced precision, the parameter updates tend to become sparser (despite stochastic rounding) due to the perilous combination of smaller gradients and diminished learning rates.~The network's performance suffers as a result and the minimum achievable test error saturates at $28.8\%$.~Fortunately, this damage is reversible as shown in Figure~\ref{CIFAR10_CNN}.~After training for $100$ epochs using the \fxpt{4}{12} format, we relax the constraint on $\tt {WL} $ slightly and increase $\tt {WL}$ by $4$ bits to $20$ bits.~This increases the fractional precision to $16$ bits (\fxpt{4}{16} format) and subsequent training results in a rapid improvement in the network's performance.~After an additional 15-20 epochs of training using the higher precision representation, the test error approaches that obtained using {\tt float}.

This result reveals a promising (and possibly more robust) strategy for deep neural network training in which the network is first trained using low-precision fixed-point arithmetic and stochastic rounding.~At the point where learning shows stagnation, the network can be \mbox{``fine-tuned"} using only a few epochs of higher-precision fixed-point computations.~Such a concept of employing mixed-precision computations has been explored previously in the context of floating point arithmetic~\cite{baboulin2009accelerating}, motivated largely by the fact that most modern processors achieve a factor $2$ to $4$ higher computational throughput for single-precision ($32$-bit) floating-point as compared with double-precision ($64$-bit) floating-point.~Similar concepts, in conjunction with stochastic rounding, can be extended to perform mixed-precision fixed-point arithmetic.\footnote{While preparing this paper, we became aware of a very recent work~\cite{courbariaux2014} that shares our motivations but adopts an orthogonal approach.~The authors propose the use of dynamic fixed-point (a hybrid of the fixed-point and  the conventional floating-point arithmetic) for training deep neural networks.~However, hardware implications of this approach are not immediately obvious.}

\section{Hardware Prototyping}
\label{HardwareSection}
The execution time of the mini-batch stochastic gradient descent algorithm  is dominated by a series of {\tt GEMM} operations in the feed-forward, error back-propagation and weight update calculation steps{\footnote{Convolution may also be rewritten as a {\tt GEMM} operation}}.~As a result, an improvement in the computational throughput of the {\tt GEMM} operation translates into an improvement in the training time.~GPUs offering a large number of parallel vector processors and high memory bandwidth have therefore been very effective in accelerating these workloads.

In this section we describe a FPGA-based hardware accelerator for matrix-matrix multiplication.~Our choice of using FPGAs as the hardware substrate is motivated by two factors.~Firstly, FPGAs enable fast hardware development times and significantly lower costs when compared to ASICs\footnote{\textbf{A}pplication \textbf{S}pecific \textbf{I}ntegrated \textbf{C}ircuits}.~Secondly, modern FPGAs have a large number of hard-wired fixed-point DSP units that are well-suited to implementing the fixed-point arithmetic described in the earlier sections, and can potentially yield gains in performance and power efficiency.~However, limited memory bandwidth must still be carefully managed through various design choices. 
\begin{figure}[ht]
%\vskip 0.2in
\begin{center}
\centerline{\includegraphics[width=\columnwidth]{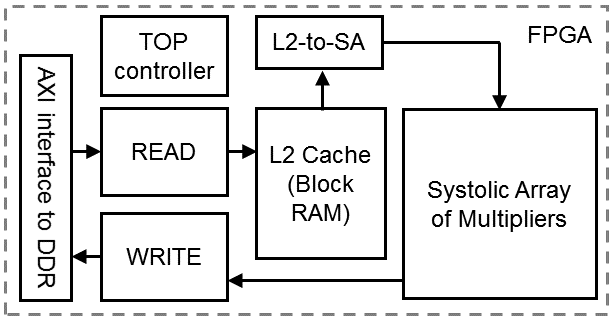}}
\caption{Block diagram of the FPGA-based fixed-point matrix multiplier.}
\label{MM_BD}
\end{center}
\vskip -0.3in
\end{figure} 

Our prototype is implemented on an off-the-shelf FPGA card featuring a
Xilinx Kintex$325$T FPGA and $8$~GB DDR$3$ memory, 
and communicating with the host PC over a PCIe bus.~This FPGA has 840 DSP multiply-accumulate units and almost $2$~MB of on-chip block RAM.~The data bandwidth between the off-chip DDR$3$ memory and the FPGA is $6.4$~GB/s.~The typical dimensions of the input matrices preclude storing entire matrices in on-chip RAM.~Thus, these matrices are stored in the DDR$3$ memory and parts of the matrices are brought into the FPGA for performing the computations.~The off-chip communication bandwidth limitation necessitates
that we reuse the on-chip data to the highest extent possible to make the
achievable throughput, measured in giga-operations/second~(G-ops/s), compute-bound.

\subsection{System Description}

Figure~\ref{MM_BD} presents a block diagram of the our fixed-point matrix multiplier. The DSP units within the FPGA are organized as a massively parallel $2$-dimensional systolic array (SA) ~\cite{HTKung} of size $n$ such that $n^2 < 840$.~This forms the core of the multiplier and will be described in greater detail in the next subsection.~Most of the block RAM on the FPGA is designated as the L$2$
cache where a fraction of the input matrices are stored.~The {\tt READ} logic
sends data requests to the DDR$3$ memory and organizes the incoming data into the L$2$ cache.~The {\tt WRITE} logic sends back computed results to the external memory.~The { \tt L$2$-to-SA} circuit moves relevant rows and columns from the L$2$ cache to the array. The {\tt TOP} controller coordinates the entire process.~The FPGA also contains Xilinx-supplied IP blocks that interface to the DDR$3$ memory. 

The operation sequence of the multiplier is as follows.~Assume the first
input matrix $A$ has dimensions $l$ x $k$ and the second input matrix $B$ has dimensions $k$ x $m$.~Initially~$n$ columns of matrix $B$ and~$pn$ rows of matrix $A$, where $p$ is the largest integer we can choose based on on-chip memory capacity constraints, are brought into the FPGA to compute $pn^{2}$ elements of the result matrix.~The next $n$ columns of matrix $B$ are then brought it and processed. This continues until all $m$ columns of matrix $B$ have been multiplied with the first $pn$ rows of matrix $A$.~This entire sequence is repeated ${l}/{pn}$ times to process
all rows of matrix $A$.~Double buffering is employed to hide the latency of
bringing in new subsets of the matrices in to the chip.~This sequence of
operation ensures that elements of matrix $A$ are reused $m$ times once
brought into the FPGA while those of matrix $B$ are reused $pn$
times.~This reuse allows efficient use of the bandwidth between the FPGA and
the DDR$3$ memory.    
\subsection{Systolic Array Architecture}

\begin{figure}[hb]
\vskip -0.1in
\begin{center}
\centerline{\includegraphics[width=3in]{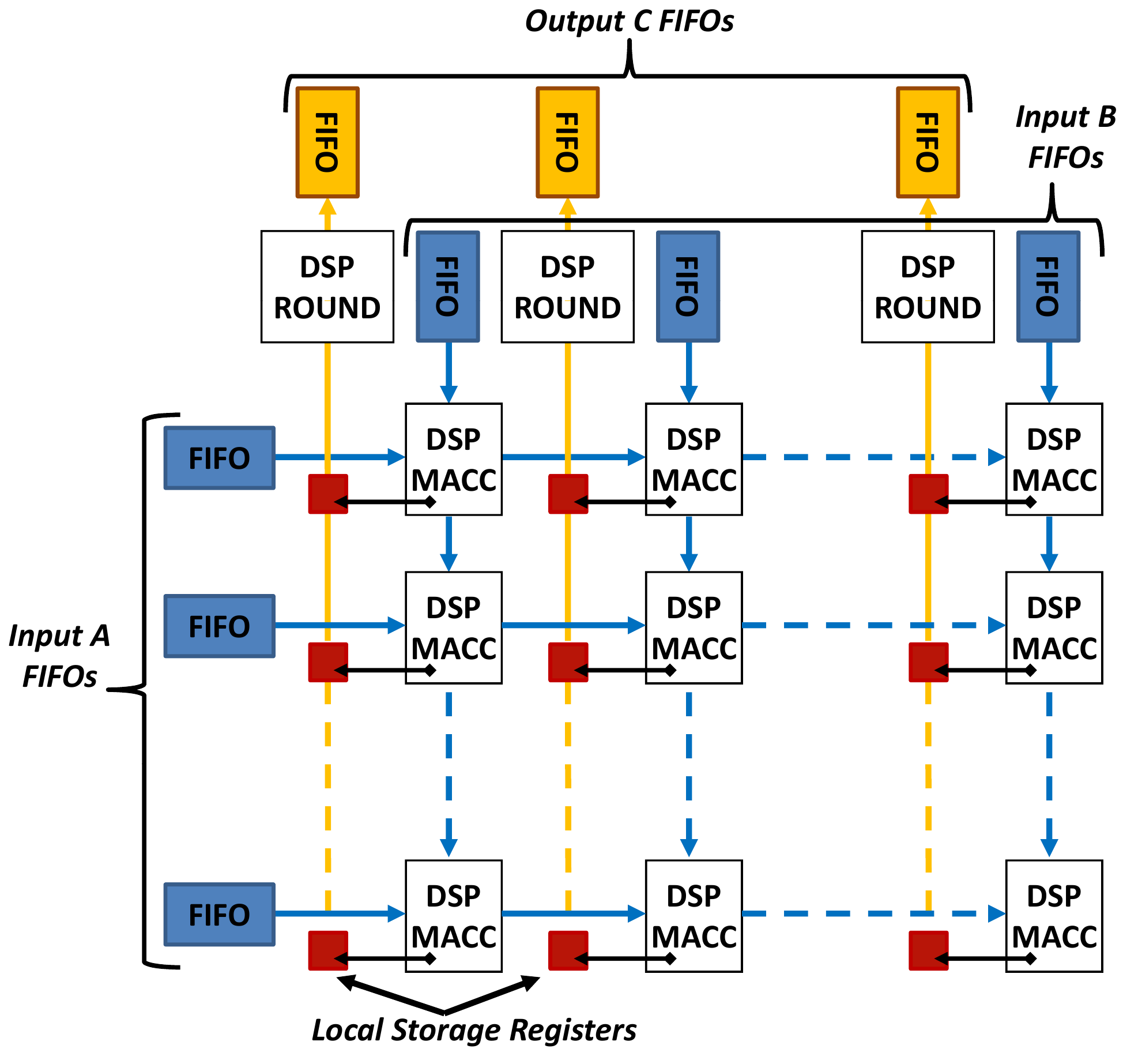}}
\vskip -0.1in
\caption{Schematic of the systolic core for matrix multiplication.}
\label{SystolicArray}
\end{center}
\vskip -0.2in
\end{figure} 

Figure~\ref{SystolicArray} shows the logical organization of 
the systolic array. Each node of the systolic array ({\tt{DSP MACC}}) has a DSP unit that implements two operations (multiply and accumulate) in every clock cycle.~Elements of input matrices $A$ and $B$ brought in from L$2$-cache are staged in local block RAM units configured as FIFO (First In First Out) queues.~Each FIFO contains elements from either a row of $A$ or a column of $B$.~
In each clock cycle, one element is read out from the FIFO.~Elements from earlier cycles are cascaded right (for $A$) or down (for $B$) and the corresponding partial products are accumulated at the DSP units.~After  accumulation of all partial products, output data is cascaded out to stochastic rounding units ({\tt{DSP ROUND}}) that are also implemented with DSP units.~Rounded results are stored in output FIFOs (one per column) before final readout to external memory.~Throughput of the array depends on the number of DSPs available and the maximum operating frequency at which the system can be operated without timing errors.~This is an example of a wavefront-type systolic array where all connections are local, i.e. only between neighboring DSPs and edge FIFOs, which limits interconnect delays and improves maximum operating frequency.

\begin{figure}[ht]
\begin{center}
\centerline{\includegraphics[width=2.8in]{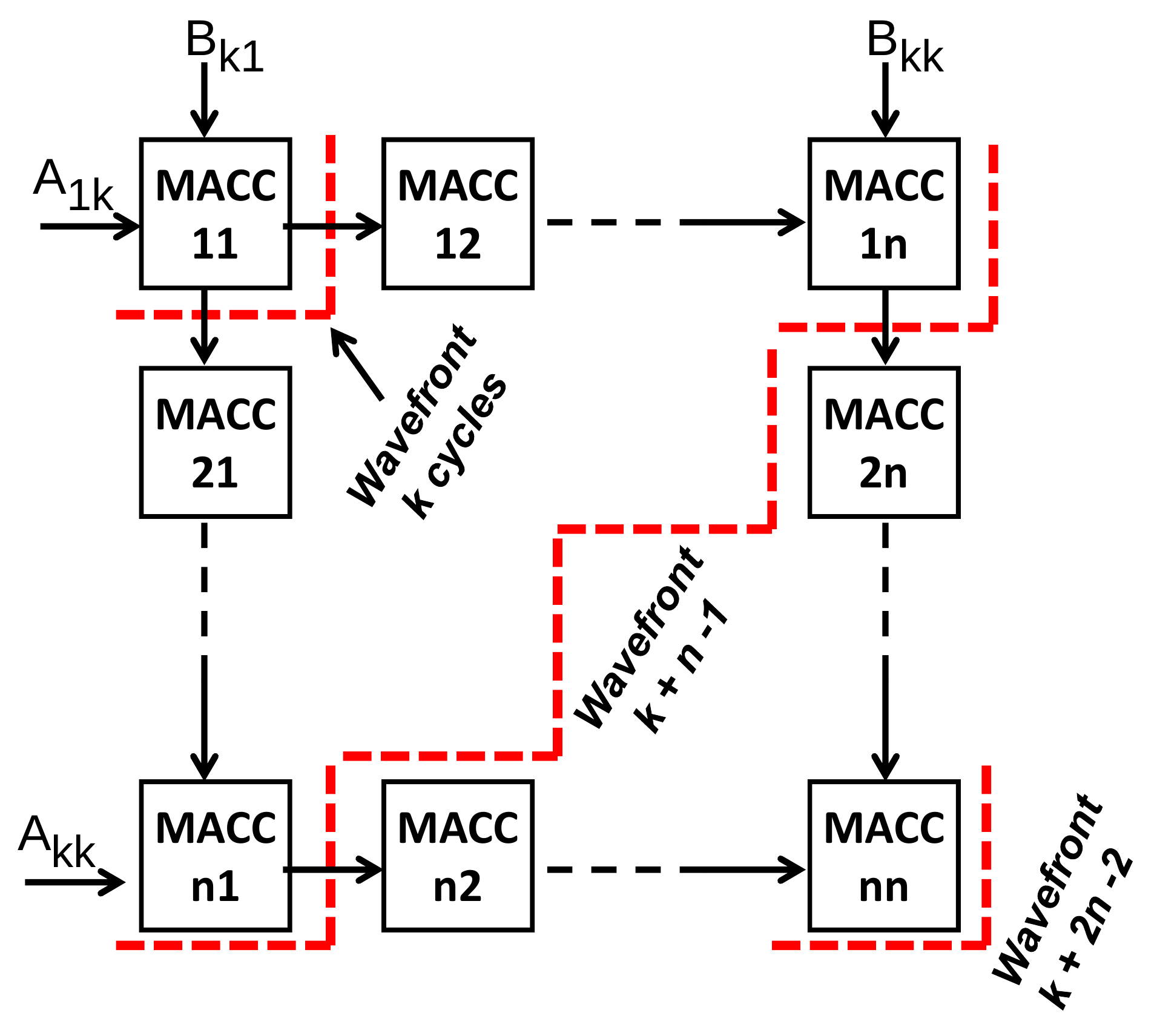}}
\vskip -0.1in
\caption{Wavefront systolic array operation.}
\label{Wavefront}
\end{center}
\vskip -0.3in
\end{figure} 

In a wavefront array, as depicted in Figure~\ref{Wavefront}, at the end of $k$ cycles, where $k$ corresponds to the inner dimension of 
the matrix multiplication, {\tt MACC} unit ``11" has accumulated
all of its partial products.~At this point, the accumulated result is 
transferred to a local register and the DSP is reset.~This 
frees it up to receive data from the next matrix multiplication operation, 
even before other elements have completed.  
This achieves high throughput for the systolic array so long as the pipeline is fed with new incoming data.~At the end of $(k+2n-2)$ cycles, the matrix multiplication is complete, and data from the last DSP unit can be read out.  
Output paths from local registers to the edge of the array are also cascaded. 

Word length of the result elements after {\tt MACC} operations are much larger (typically $48$ bits if using $7$-series DSPs) than word length of the inputs (typically $18$ bits or less).~Before transferring to output FIFOs, result elements must be trimmed through the 
stochastic rounding of least signficant bits (LSB) and truncation of excess MSB bits (after detection of overflow/underflow).~Both operations can 
be efficiently achieved using a single DSP unit per output.~At each column, linear feedback shift register (LFSR) is used 
to generate a random number whose width is equal to the number of LSB bits being 
rounded off. The DSP unit adds the random number 
to the incoming result and drops rounded off LSB bits. 
Pattern-detect capabilities built into the DSP 
are used to determine if excess MSB bits are identical 
(all ``$0$s" or all ``$1$s").~If not, an overflow/underflow condition is detected, 
and result values are saturated to the max/min $2$'s complement values\footnote
{A more direct stochastic rounding approach is 
multi-bit magnitude comparison of result LSB vs. a 
random number, followed by a conditional addition and examining excess MSBs.~The approach in this section achieves the same result but removes the first full multi-bit
comparison, enabling compact implementation on a single DSP unit. }.~The result is then 
transferred to output column FIFOs awaiting writeback to external memory.~The overhead of stochastic rounding is thus the logic occupied by {\tt{DSP ROUND}} units, which in our case is $28$ DSP units -- corresponding to less than $4\%$ overhead in hardware resources.

\subsection{Results}
For a $28$x$28$ systolic array implemented on the KintexK$325$T FPGA, Xilinx's Vivado synthesis and place-and-route tool estimated a maximum 
circuit operation frequency of $166$~MHz and a power consumption of $7$~W.~This translates to a throughput of $260$~G-ops/s at a power efficiency of $37$~G-ops/s/W.~This compares very favorably against the Intel i$7$-$3720$QM CPU, the NVIDIA GT$650$m and the GTX$780$ GPUs, all of which achieve power efficiency in the range of $1$-$5$ G-ops/s/W~\cite{gokhale2014240}.~Table~\ref{fpga-usage} presents a summary of the utilization of various resources in the FPGA.~Throughput numbers can benefit from migration to newer Xilinx FPGAs, 
such as the Ultrascale series, that have much higher number of DSP units and can potentially operate at higher frequencies. 
\begin{table}[h]
\vskip -0.1in
\caption{FPGA resource utilization.}
\begin{center}
\begin{small}
\begin{sc}
\begin{tabular}{lcccr}
\hline
\abovespace\belowspace
Resource & \multicolumn{1}{l}{Usage} & \multicolumn{1}{l}{\begin{tabular}[c]{@{}l@{}}Available on \\ XCVK$325$T\end{tabular}} & \begin{tabular}[c]{@{}c@{}}Utilization \\ Ratio\end{tabular} \\
\hline
\abovespace
LUTs       & 62922                     & 203800                                                                               & 31\%                                                         \\
Flip-flops & 146510                    & 407600                                                                               & 36\%                                                         \\
DSP        & 812                       & 840                                                                                  & 97\%                                                         \\
\belowspace
Block RAM  & 334                       & 445                                                                                  & 75\%  \\

\hline
\end{tabular}
\end{sc}
\end{small}
\end{center}
\vskip -0.1in
\label{fpga-usage}
\end{table}

\section{Conclusion}

In this paper, we embrace a top-down approach exploiting the noise-tolerance of deep neural networks and their training algorithms to influence the design of low-level compute units.~Specifically, the substitution of floating-point units with fixed-point arithmetic circuits comes with significant gains in the energy efficiency and computational throughput, while potentially risking the neural network's performance.~For low-precision fixed-point computations, where conventional rounding schemes fail, adopting stochastic rounding during deep neural network training delivers results nearly identical as 32-bit floating-point computations.~Additionally, we implement a high-throughput, energy-efficient architecture for matrix multiplication that incorporates stochastic rounding with very little overhead.~Extrapolating, we envision the emergence of hardware-software co-designed systems for large-scale machine learning based on relaxed, inexact models of computing running on non-deterministic components all across the stack, right down to low-level hardware circuitry. 

\bibliography{ICML_arxiv}
\bibliographystyle{icml2015}

\end{document}